\pdfoutput=1
\documentclass[11pt,a4paper]{article}
\usepackage{naaclhlt2019}
\usepackage[T1]{fontenc}
\usepackage{times}
\usepackage{url}
\usepackage{latexsym}
\usepackage{amsmath}
\usepackage{breqn}
\usepackage{pgfplotstable}
\usepackage{hhline}
\usepackage{multirow}
\usepackage{multicol}
\usepackage{caption}
\usepackage{subcaption}
\usepackage{color}
\usepackage{float}
\usepackage{lipsum,adjustbox}
\usepackage{tikz}
\usepackage{tikz-dependency}
\usepackage{rotating}
\usepackage{listings}
\usetikzlibrary{shapes,fit,calc,er,positioning,intersections,decorations.shapes,mindmap,trees}
\usepackage{enumitem}
\tikzset{decorate sep/.style 2 args={decorate,decoration={shape backgrounds,shape=circle,
      shape size=#1,shape sep=#2}}}

\newcommand{\com}[1]{}

\aclfinalcopy

\usepackage{color}
\definecolor{gray}{rgb}{0.4,0.4,0.4}
\definecolor{darkblue}{rgb}{0.0,0.0,0.6}
\definecolor{cyan}{rgb}{0.0,0.6,0.6}

\lstdefinelanguage{XML}
{
  morestring=[b]",
  morestring=[s]{>}{<},
  morecomment=[s]{<?}{?>},
  stringstyle=\color{black},
  identifierstyle=\color{darkblue},
  keywordstyle=\color{cyan},
  morekeywords={xmlns,version,type}
}

\title{SemEval-2019 Task 1: \\ Cross-lingual Semantic Parsing with UCCA}

\author{Daniel Hershcovich \\  {\bf Elior Sulem} \And
  Zohar Aizenbud \\ {\bf Ari Rappoport} \\
  School of Computer Science and Engineering \\
  Hebrew University of Jerusalem \\
  \texttt{\{danielh,zohara,borgr,eliors,arir,oabend\}@cs.huji.ac.il} \\
  \And Leshem Choshen \\ {\bf Omri Abend}
}

\begin{document}
\maketitle

\begin{abstract}
  We present the SemEval 2019 shared task on
  Universal Conceptual Cognitive Annotation
  (UCCA) parsing in English, German and French, 
  and discuss the participating systems and results.
  UCCA is a cross-linguistically applicable framework for semantic
  representation, which builds on extensive typological work and supports rapid annotation.
  UCCA poses a challenge for existing parsing techniques,
  as it exhibits reentrancy (resulting in DAG structures),
  discontinuous structures and non-terminal nodes corresponding
  to complex semantic units.
  The shared task has yielded improvements over the state-of-the-art 
  baseline in all languages and settings. Full results can be found
  in the task's website  \url{https://competitions.codalab.org/competitions/19160}.
\end{abstract}

\section{Overview}

  Semantic representation is receiving growing attention in NLP in the past few years,
  and many proposals for semantic schemes have recently been put forth. Examples
  include Abstract Meaning Representation \cite[AMR;][]{banarescu2013abstract},
  Broad-coverage Semantic Dependencies \cite[SDP;][]{oepen2016towards},
  Universal Decompositional Semantics \cite[UDS;][]{white2016universal},
  Parallel Meaning Bank \cite{abzianidze2017the},
  and Universal Conceptual Cognitive Annotation \cite[UCCA;][]{abend2013universal}.
  These advances in semantic representation, along with corresponding advances in semantic parsing,
  can potentially benefit essentially all text understanding tasks, and have already demonstrated applicability to
  a variety of tasks, including summarization \citep{liu2015toward,Dohare2017TextSU}, 
  paraphrase detection \cite{issa-18},  and semantic evaluation (using UCCA; see below).
In this shared task, we focus on UCCA parsing in multiple languages.
One of our goals is to benefit semantic parsing in languages with less
annotated resources by making use of data from more resource-rich languages.
We refer to this approach as \textit{cross-lingual} parsing,
while other works \cite{I17-1084,D18-1194} define cross-lingual parsing
as the task of parsing text in one language to meaning representation in another
language.

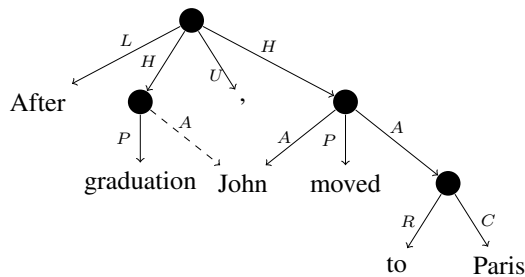
\begin{figure}[t]
  \centering
  \scalebox{.9}{
  \begin{tikzpicture}[level distance=12mm, ->]
    \node (ROOT) [fill=black, circle] {}
      child {node (After) {After} edge from parent node[above] {\scriptsize $L$}}
      child {node (graduation) [fill=black, circle] {}
      {
        child {node {graduation} edge from parent node[left] {\scriptsize $P$}}
      } edge from parent node[left] {\scriptsize $H$} }
      child {node {,} edge from parent node[below] {\scriptsize $U$}}
      child {node (moved) [fill=black, circle] {}
      {
        child {node (John) {John} edge from parent node[left] {\scriptsize $A$}}
        child {node {moved} edge from parent node[left] {\scriptsize $P$}}
        child {node [fill=black, circle] {}
        {
          child {node {to} edge from parent node[left] {\scriptsize $R$}}
          child {node {Paris} edge from parent node[right] {\scriptsize $C$}}
        } edge from parent node[above] {\scriptsize $A$} }
      } edge from parent node[above] {\scriptsize $H$} }
      ;
    \draw[dashed,->] (graduation) to node [above] {\scriptsize $A$} (John);
  \end{tikzpicture}}
\caption{\label{fig:example}
  An example UCCA graph.}
\end{figure}

\begin{table*}[t]
\centering
\footnotesize
\begin{tabular}{cp{2cm}p{125mm}}
\multicolumn{3}{c}{\bf Scene Elements}\\
P & {\bf Process} & The main relation of a Scene that evolves in time (usually an action or movement).\\
S & {\bf State} & The main relation of a Scene that does not evolve in time.\\
A & {\bf Participant} & Scene participant (including locations, abstract entities and Scenes serving as arguments).\\
D & {\bf Adverbial} & A secondary relation in a Scene.\\
\multicolumn{3}{c}{{\bf Elements of Non-Scene Units}}\\
C & {\bf Center} & Necessary for the conceptualization of the parent unit.\\
E & {\bf Elaborator} & A non-Scene relation applying to a single Center.\\
N & {\bf Connector} & A non-Scene relation applying to two or more Centers, highlighting a common feature.\\
R & {\bf Relator} & All other types of non-Scene relations: (1) Rs that relate a C to some super-ordinate relation, and
(2) Rs that relate two Cs pertaining to different aspects of the parent unit. \\
\multicolumn{3}{c}{\bf Inter-Scene Relations}\\
H & {\bf Parallel Scene} & A Scene linked to other Scenes by regular linkage (e.g., temporal, logical, purposive).\\
L & {\bf Linker} & A relation between two or more Hs (e.g., ``when'', ``if'', ``in order to'').\\
G & {\bf Ground} & A relation between the speech event and the uttered Scene (e.g., ``surprisingly'').\\
\multicolumn{3}{c}{\bf Other}\\
F & {\bf Function} & Does not introduce a relation or participant. Required by some structural pattern.
\end{tabular}
\caption{The complete set of categories in UCCA's foundational layer.\label{tab:ucca}}
\end{table*}

  In addition to its potential applicative value, work on
  semantic parsing poses interesting algorithmic and modeling challenges,
  which are often different from those tackled in syntactic parsing, 
  including reentrancy (e.g., for sharing arguments across predicates),
  and the modeling of the interface with lexical semantics.
  
  UCCA is a cross-linguistically applicable semantic representation scheme,
  building on the established Basic Linguistic Theory typological framework \citep{Dixon:10a,Dixon:10b,Dixon:12}.
  It has demonstrated applicability to multiple languages, including
  English, French and German, and pilot annotation projects were conducted on a few languages more.
  UCCA structures have been shown to be well-preserved in translation \citep{sulem2015conceptual},
  and to support rapid annotation by non-experts, assisted by an accessible annotation interface
  \citep{abend2017uccaapp}.\footnote{\url{https://github.com/omriabnd/UCCA-App}}
  UCCA has already shown applicative value for text simplification \citep{sulem2018simple},  
  as well as for defining semantic evaluation measures for text-to-text generation tasks,
  including machine translation \cite{birch2016hume}, text simplification \cite{sulem2018semantic} and grammatical error correction
  \cite{choshen2018reference}.
  
  The shared task defines a number of tracks, based on the different corpora and the availability of external resources (see \S\ref{sec:evaluation}).  
  It received submissions from eight research groups around the world.
  In all settings at least one of the submitted systems improved over the state-of-the-art TUPA parser \citep{hershcovich2017a,hershcovich2018multitask}, used
  as a baseline.

\section{Task Definition}\label{sec:task_definition}

  UCCA represents the semantics of linguistic utterances
  as directed acyclic graphs (DAGs), where terminal (childless) nodes
  correspond to the text tokens, and non-terminal nodes to semantic
  units that participate in some super-ordinate relation.
  Edges are labeled, indicating the role of a child in the relation the
  parent represents. Nodes and edges belong to one of several
  \textit{layers}, each corresponding to a ``module'' of
  semantic distinctions.

  UCCA's \textit{foundational layer} covers the predicate-argument
  structure evoked by predicates of all grammatical categories
  (verbal, nominal, adjectival and others), the inter-relations between them,
  and other major linguistic phenomena such as semantic heads and multi-word expressions.
  It is the only layer for which annotated corpora exist at the moment, and is thus the target of this shared task.
  The layer's basic notion is the \textit{Scene},
  describing a state, action, movement or some other relation that evolves in time.
  Each Scene contains one main relation (marked as either a Process or a State),
  as well as one or more Participants.
  For example, the sentence ``After graduation, John moved to Paris'' (Figure~\ref{fig:example})
  contains two Scenes, whose main relations are ``graduation'' and ``moved''.
  ``John'' is a Participant in both Scenes, while ``Paris'' only in the latter.
  Further categories account for inter-Scene relations and the internal structure of
  complex arguments and relations (e.g., coordination and multi-word expressions).
Table~\ref{tab:ucca} provides a concise description of the
categories used by the UCCA foundational layer.

  UCCA distinguishes \textit{primary} edges, corresponding
  to explicit relations, from \textit{remote} edges (appear dashed in
  Figure~\ref{fig:example}) that allow for a unit to participate
  in several super-ordinate relations.
  Primary edges form a tree in each layer, whereas remote edges enable reentrancy, forming a DAG.

\begin{figure*}[t]
  \centering
  \scalebox{.7}{
  \begin{tikzpicture}[level distance=29mm, ->,
  level 1/.style={sibling distance=6em},
  level 2/.style={sibling distance=3em},
  level 3/.style={sibling distance=4.3em},
  every node/.append style={anchor=west,text height=.6ex,text depth=0}]
    \node (ROOT) [fill=black, circle] {}
      child {node [fill=black, circle] {}
      {
        child {node {A} edge from parent node[left] {$E$}}
        child {node {similar} edge from parent node[left] {$E$}}
        child {node {technique} edge from parent node[right] {$C$}}
      } edge from parent node[left] {$A\quad$ \hspace{1mm} } }
      child {node {is} edge from parent node[left] {$F$}}
      child {node [fill=black, circle] {}
      {
        child {node {almost} edge from parent node[left] {$E$}}
        child {node {impossible} edge from parent node[right] {$C$}}
      } edge from parent node[left] {$D$} }
      child {node {\textbf{IMPLICIT}} edge from parent node[left] {$A$}}
      child {node {to} edge from parent node[left] {$F$}}
      child {node {apply} edge from parent node[left] {$P\quad$}}
      child {node [fill=black, circle] {}
      {
        child {node {to} edge from parent node[left] {$R$}}
        child {node {other} edge from parent node[left] {$E$}}
        child {node {crops} edge from parent node[left] {$C$}}
        child {node {,} edge from parent node[left] {$U$}}
        child {node [fill=black, circle] {}
        {
          child {node {such as} edge from parent node[left] {$R$}}
          child {node {cotton} edge from parent node[left] {$C$}}
          child {node {,} edge from parent node[left] {$U$}}
          child {node {soybeans} edge from parent node[left] {$C$}}
          child {node {and} edge from parent node[left] {$N$}}
          child {node {rice} edge from parent node[right] {$\; C$}}
        } edge from parent node[right] {$\; E$ \hspace{1mm} } }
      } edge from parent node[left] {$A\;$ \hspace{1mm} } }
      child {node {.} edge from parent node[right] {$\quad \quad U$}}
      ;
  \end{tikzpicture}
  }
  \caption{UCCA example with an implicit unit.}
  \label{fig:example_implicit}
\end{figure*}
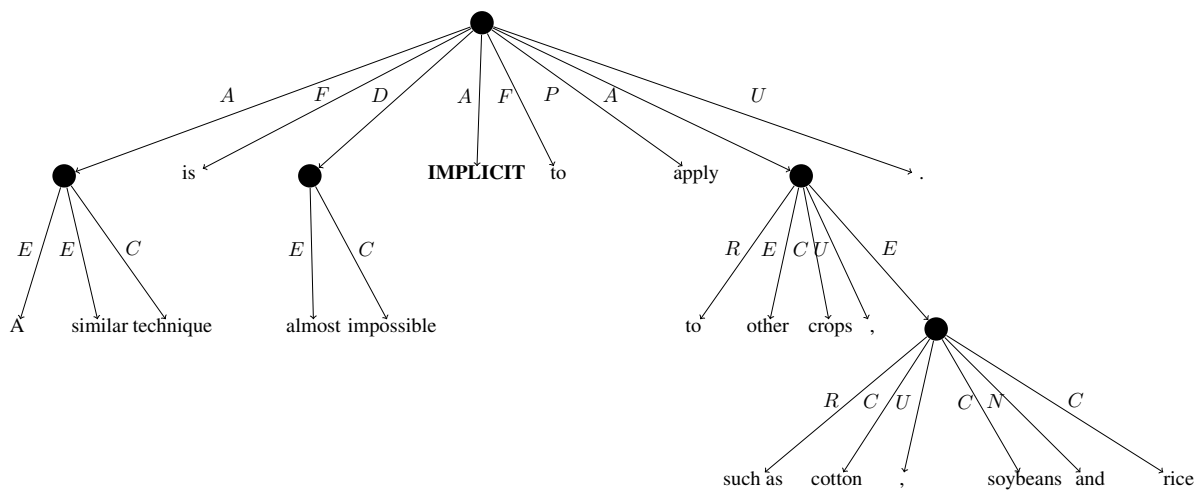

UCCA graphs may contain \textit{implicit} units with no correspondent in the text.
Figure~\ref{fig:example_implicit} shows the annotation for the sentence
``A similar technique is almost impossible to apply to other crops, such as cotton, soybeans
and rice.''\footnote{The same example was used by \citet{oepen2015semeval} to compare different semantic dependency schemes.}
It includes a single Scene, whose main relation is ``apply'', a secondary relation ``almost impossible'', as well as two complex arguments: ``a similar technique'' and the coordinated argument ``such as cotton, soybeans, and rice.''
In addition, the Scene includes an implicit argument, which represents the agent of the
``apply'' relation.

While parsing technology is well-established for syntactic parsing,
UCCA has several formal properties
that distinguish it from syntactic representations,
mostly UCCA's tendency to abstract away from syntactic detail that do not
affect argument structure.
For instance, consider the following examples where the concept of a Scene
has a different rationale from the syntactic concept of a clause.
First, non-verbal predicates in UCCA are represented like verbal ones,
such as when they appear in copula clauses or noun phrases. Indeed,
in Figure \ref{fig:example}, ``graduation'' and ``moved'' are considered separate Scenes,
despite appearing in the same clause.
Second, in the same example, ``John'' is marked as a (remote) Participant
in the graduation Scene, despite not being explicitly mentioned.
Third, consider the possessive construction in ``John's trip home''.
While in UCCA ``trip'' evokes a Scene in which ``John'' is
a Participant, a syntactic scheme would analyze this phrase similarly to ``John's shoes''.

The differences in the challenges posed by syntactic parsing and UCCA parsing, and more generally by semantic parsing,
motivate the development of targeted parsing technology to tackle it.

\section{Data \& Resources}

  All UCCA corpora are freely available.\footnote{\url{https://github.com/UniversalConceptualCognitiveAnnotation}}
  For English, we use v1.2.3 of the Wikipedia UCCA corpus ({\it Wiki}), v1.2.2 of the UCCA {\it Twenty Thousand Leagues Under the Sea}
  English-French parallel corpus ({\it 20K}),
  which includes UCCA manual annotation for the first five chapters in French and English,
  and v1.0.1 of the UCCA German {\it Twenty Thousand Leagues Under the Sea} corpus,
  which includes the entire book in German.
  For consistent annotation, we replace any Time and Quantifier labels with 
  Adverbial and Elaborator in these data sets.
  The resulting training,
  development\footnote{\scriptsize\url{http://bit.ly/semeval2019task1traindev}}
  and test sets\footnote{\scriptsize\url{http://bit.ly/semeval2019task1test}}
  are publicly available, and the splits are given in Table~\ref{table:data_split}.
  Statistics on various structural properties are given in Table~\ref{table:data}.

  \begin{table*}[t]
      \small
      \setlength\tabcolsep{5pt}
      \centering
      \begin{tabular}{l|c|c|c|c|c|c|c|c|c}
          & \multicolumn{2}{c|}{train/trial} & \multicolumn{2}{c|}{dev} & \multicolumn{2}{c|}{test} & \multicolumn{3}{c}{total} \\
          corpus             & sentences & tokens & sentences & tokens & sentences & tokens & passages & sentences & tokens \\
          \hline
          English-Wiki & 4,113      & 124,935 & 514       & 17,784  & 515       & 15,854  & 367      & 5,142      & 158,573 \\
          English-20K  & 0         & 0      & 0         & 0      & 492       & 12,574  & 154      & 492       & 12,574  \\
          French-20K   & 15        & 618    & 238       & 6,374   & 239       & 5,962   & 154      & 492       & 12,954  \\
          German-20K   & 5,211      & 119,872 & 651       & 12,334  & 652       & 12,325  & 367      & 6,514      & 144,531
      \end{tabular}
      \caption{Data splits of the corpora used for the shared task.\label{table:data_split}}
  \end{table*}

  \begin{table}[ht]
  \centering
  \setlength\tabcolsep{2.5pt}
  \small
\begin{tabular}{l|rrrr}
    & En-Wiki & En-20K & Fr-20K & De-20K \\
    \hline
    \# passages&367&154&154&367 \\
    \# sentences&5,141&492&492&6,514 \\
    \# tokens&158,739&12,638&13,021&144,529 \\
    \hline
    \# {non-terminals}&62,002&4,699&5,110&51,934 \\
    \% {discontinuous}&1.71&3.19&4.64&8.87 \\
    \% {reentrant}&1.84&0.89&0.65&0.31 \\
    \hline
    \# edges&208,937&16,803&17,520&187,533 \\
    \% primary&97.40&96.79&97.02&97.32 \\
    \% remote&2.60&3.21&2.98&2.68\\
    by category &\\
\;\% Participant&17.17&18.1&17.08&19.86\\
\;\% Center&18.74&16.31&18.03&14.32\\
\;\% Adverbial&3.65&5.25&4.18&5.67\\
\;\% Elaborator&18.98&18.06&18.65&14.88\\
\;\% Function&3.38&3.58&2.58&2.98\\
\;\% Ground&0.03&0.56&0.37&0.57\\
\;\% Parallel Scene&6.02&6.3&6.15&7.54\\
\;\% Linker&2.19&2.66&2.57&2.49\\
\;\% Connector&1.26&0.93&0.84&0.65\\
\;\% Process&7.1&7.51&6.91&7.03\\
\;\% Relator&8.58&8.09&9.6&7.54\\
\;\% State&1.62&2.1&1.88&3.34\\
\;\% Punctuation&11.28&10.55&11.16&13.15\\
\end{tabular}
      \caption{Statistics of the corpora used for the shared task.\label{table:data}}
  \end{table}

  The corpora were manually annotated according to v1.2 of the UCCA
  guidelines,\footnote{\scriptsize\url{http://bit.ly/semeval2019task1guidelines}}
  and reviewed by a second annotator. All data was passed through automatic validation 
  and normalization scripts.\footnote{\url{https://github.com/huji-nlp/ucca/tree/master/scripts}}
  The goal of validation is to rule out cases that are inconsistent with the UCCA annotation guidelines. 
  For example, a Scene, defined by the presence of a Process or a State, should include at least one Participant.

  Due to the small amount of annotated data available for French, we only provided a minimal training set
  of 15 sentences, in addition to the development and test set.
  Systems for French were expected to pursue semi-supervised approaches, such as 
  cross-lingual learning or structure projection, leveraging the parallel nature of the corpus,
  or to rely on datasets for related formalisms, such as Universal Dependencies \cite{nivre2016universal}.
  The full unannotated 20K Leagues corpus in English and French was released as well, 
  in order to facilitate pursuing cross-lingual approaches.

  Datasets were released in an XML format, including tokenized text
  automatically pre-processed using spaCy (see \S\ref{sec:evaluation}),
  and gold-standard UCCA annotation for the train and development sets.\footnote{\url{https://github.com/UniversalConceptualCognitiveAnnotation/docs/blob/master/FORMAT.md}}
  To facilitate the use of existing NLP tools, we also released the data in SDP, AMR, CoNLL-U and plain text formats.

\section{TUPA: The Baseline Parser}\label{sec:pilot}

    We use the TUPA parser, the only parser for UCCA at the time the task was announced, as a baseline \citep{hershcovich2017a,hershcovich2018multitask}.
    TUPA is a transition-based DAG parser based on a BiLSTM-based classifier.\footnote{\url{https://github.com/huji-nlp/tupa}}
    TUPA in itself has been found superior to a number of conversion-based parsers that use existing parsers for other formalisms 
    to parse UCCA by constructing a two-way conversion protocol between the formalisms.
    It can thus be regarded as a strong baseline for system submissions to the shared task.

\section{Evaluation}\label{sec:evaluation}

\paragraph{Tracks.}
  Participants in the task were evaluated in four settings:

  \begin{enumerate}
    \item
      English in-domain setting, using the Wiki corpus.
    \item
      English out-of-domain setting, using the Wiki corpus as training and development data, and 20K Leagues as test data.
    \item
      German in-domain setting, using the 20K Leagues corpus.
    \item
      French setting with no training data, using the 20K Leagues as development and test data.
  \end{enumerate}

  In order to allow both even ground comparison between systems and using hitherto untried resources,
  we held both an \textbf{open} and a \textbf{closed} track for submissions in the English and German settings.
  Closed track submissions were only allowed to use the gold-standard UCCA annotation
  distributed for the task in the target language, and were limited in their use of additional resources.
  Concretely, the only additional data they were allowed to use is that used by TUPA,
  which consists of automatic annotations provided by spaCy:\footnote{\url{http://spacy.io}}
  POS tags, syntactic dependency relations, and named entity types and spans.
  In addition, the closed track only allowed the use of word embeddings provided by fastText
  \cite{bojanowski2016enriching}\footnote{\url{http://fasttext.cc}} for all languages.

  Systems in the open track, on the other hand, were allowed to use any additional resource,
  such as UCCA annotation in other languages, dictionaries or datasets for other tasks,
  provided that they make sure not to use any additional gold standard annotation over
  the same text used in the UCCA corpora.\footnote{We are not aware of any such annotation, but
    include this restriction for completeness.}
  In both tracks, we required that submitted systems are not trained on the development data.
  We only held an open track for French, due to the paucity of training data. 
  The four settings and two tracks result in a total of 7 competitions.

\paragraph{Scoring.}
  The following scores an output graph $G_1=(V_1,E_1)$ against a gold one,
  $G_2=(V_2,E_2)$, over the same sequence of terminals (tokens) $W$.
  For a node $v$ in $V_1$ or $V_2$, define $yield(v) \subseteq W$ as is its
  set of terminal descendants.
  A pair of edges $(v_1,u_1) \in E_1$ and $(v_2,u_2) \in E_2$ with labels
  (categories) $\ell_1, \ell_2$ is \textit{matching}
  if $yield(u_1) = yield(u_2)$ and $\ell_1=\ell_2$.
  Labeled precision and recall are defined by dividing the number of matching edges
  in $G_1$ and $G_2$ by $|E_1|$ and $|E_2|$, respectively.
  $F_1$ is their harmonic mean:
  \[2\cdot\frac{\mathrm{Precision}\cdot\mathrm{Recall}}{\mathrm{Precision}+\mathrm{Recall}}\]
  
  Unlabeled precision, recall and $F_1$ are the same,
  but without requiring that $\ell_1=\ell_2$ for the edges to match.
  We evaluate these measures for primary and remote edges separately.
  For a more fine-grained evaluation, we additionally report precision, recall and $F_1$
  on edges of each category.\footnote{The official evaluation script providing both coarse-grained and fine-grained scores can be found in
  \url{https://github.com/huji-nlp/ucca/blob/master/scripts/evaluate_standard.py}.}

\section{Participating Systems}\label{sec:participating_systems}

We received a total of eight submissions to the different tracks:
{\it ​MaskParse@Deski\~{n}} \citep{Marzinotto2019} from Orange Labs and Aix-Marseille University,
{\it HLT@SUDA} \citep{Jiang2019} from Soochow University,
{\it T\"{u}Pa} \citep{PutzGlocker2019} from the University of T\"{u}bingen,
{\it UC Davis} \citep{YuSagae2019} from the University of California, Davis ,
{\it GCN-Sem}  \citep{Taslimipoor2019} from the University of Wolverhampton,
{\it CUNY-PekingU} \citep{Lyu2019} from	the City University of New York and Peking University,
{\it DANGNT@UIT.VNU-HCM} \citep{NguyenTran2019} from the University of Information Technology VNU-HCM,
and {\it XLangMo} from Zhejiang University.
Two systems ({\it HLT@SUDA} and {\it CUNY-PekingU}) participated in all the tracks.\footnote{It was later discovered that {\it CUNY-PekingU} used some of the evaluation data for training in the open tracks, and they were thus disqualified from these tracks.}

In terms of parsing approaches, the task was quite varied. 
{\it HLT@SUDA} converted UCCA graphs to constituency trees and trained a constituency parser and a recovery mechanism of remote edges in a multi-task framework.
{\it ​MaskParse@Deski\~{n}} used a bidirectional GRU tagger with a masking mechanism. 
{\it T\"{u}pa} and {\it XLangMo} used a transition-based approach.
{\it UC Davis} used an encoder-decoder architecture.
{\it GCN-SEM} uses a BiLSTM model to predict Semantic Dependency Parsing tags, when the syntactic dependency tree is given in the input.
{\it CUNY-PKU} is based on an ensemble that includes different variations of the TUPA parser.
{\it DANGNT@UIT.VNU-HCM} converted syntactic dependency trees to UCCA graphs. 

Different systems handled remote edges differently. 
{\it DANGNT@UIT.VNU-HCM} and {\it GCN-SEM} ignored remote edges. {\it UC Davis} used a different BiLSTM for remote edges.
{\it HLT@SUDA} marked remote edges when converting the graph to a constituency tree and trained a classification model for their recovery.
{\it ​MaskParse@Deski\~{n}} handles remote edges by detecting arguments that are outside of the parent's node span using a detection threshold on the output probabilities.

In terms of using the data, all teams but one used the UCCA XML format, two used the CoNLL-U format, which is derived
by a lossy conversion process, and only one team found the other data formats helpful. One of the teams ({\it ​MaskParse@Deski\~{n}}) built a new training data adapted to their model by repeating each sentence N times, N being the number of non-terminal nodes in the UCCA graphs.
Three of the teams adapted the baseline TUPA parser, or parts of it to form their parser, namely {\it T\"{u}Pa}, 
{\it CUNY-PekingU} and {\it XLangMo}; {\it HLT@SUDA} used a constituency parser \citep{Stern2017} as a component in their model; {\it DANGNT@UIT.VNU-HCM} is a rule-based system over the Stanford Parser, and the rest are newly constructed parsers. 
\begin{table}[t]
\tiny
\setlength\tabcolsep{4.6pt}
\begin{tabular}{rlrrrrrr}
&& \multicolumn{3}{c}{Labeled} & \multicolumn{3}{c}{Unlabeled} \\
\# & Team & All & Prim. & Rem. & All & Prim. & Rem. \\
\hline
\multicolumn{8}{l}{English-Wiki (closed)} \\
1 & HLT@SUDA & 77.4 & 77.9 & 52.2 & 87.2 & 87.9 & 52.5\\
2 & baseline & 72.8 & 73.3 & 47.2 & 85.0 & 85.8 & 48.4\\
3 & Davis & 72.2 & 73.0 & 0 & 85.5 & 86.4 & 0\\
4 & CUNY-PekingU & 71.8 & 72.3 & 49.5 & 84.5 & 85.2 & 50.1\\
5 & \multirow{2}{19mm}{DANGNT@UIT.\\VNU-HCM} & 70.0 & 70.7 & 0 & 81.7 & 82.6 & 0\\\\
6 & GCN-Sem & 65.7 & 66.4 & 0 & 80.9 & 81.8 & 0\\
\multicolumn{8}{l}{English-Wiki (open)} \\
1 & HLT@SUDA & 80.5 & 81.0 & 58.8 & 89.7 & 90.3 & 60.7\\
2 & baseline & 73.5 & 73.9 & 53.5 & 85.1 & 85.7 & 54.3\\
3 & T\"uPa & 73.5 & 74.1 & 42.5 & 85.3 & 86.2 & 43.1\\
4 & XLangMo & 73.1 & 73.5 & 53.2 & 85.1 & 85.7 & 53.5\\
5 & \multirow{2}{19mm}{DANGNT@UIT.\\VNU-HCM} & 70.3 & 71.1 & 0 & 81.7 & 82.6 & 0\\\\
\multicolumn{8}{l}{English-20K (closed)} \\
1 & HLT@SUDA & 72.7 & 73.6 & 31.2 & 85.2 & 86.4 & 32.1\\
2 & baseline & 67.2 & 68.2 & 23.7 & 82.2 & 83.5 & 24.3\\
3 & CUNY-PekingU & 66.9 & 67.9 & 27.9 & 82.3 & 83.6 & 29.0\\
4 & GCN-Sem & 62.6 & 63.7 & 0 & 80.0 & 81.4 & 0\\
\multicolumn{8}{l}{English-20K (open)} \\
1 & HLT@SUDA & 76.7 & 77.7 & 39.2 & 88.0 & 89.2 & 41.4\\
2 & T\"uPa & 70.9 & 71.9 & 29.6 & 84.4 & 85.7 & 30.7\\
3 & XLangMo & 69.5 & 70.4 & 36.6 & 83.5 & 84.6 & 38.5\\
4 & baseline & 68.4 & 69.4 & 25.9 & 82.5 & 83.9 & 26.2\\
\multicolumn{8}{l}{German-20K (closed)} \\
1 & HLT@SUDA & 83.2 & 83.8 & 59.2 & 92.0 & 92.6 & 60.9\\
2 & CUNY-PekingU & 79.7 & 80.2 & 59.3 & 90.2 & 90.9 & 59.9\\
3 & baseline & 73.1 & 73.6 & 47.8 & 85.9 & 86.7 & 48.2\\
4 & GCN-Sem & 71.0 & 72.0 & 0 & 85.1 & 86.2 & 0\\
\multicolumn{8}{l}{German-20K (open)} \\
1 & HLT@SUDA & 84.9 & 85.4 & 64.1 & 92.8 & 93.4 & 64.7\\
2 & baseline & 79.1 & 79.6 & 59.9 & 90.3 & 91.0 & 60.5\\
3 & T\"uPa & 78.1 & 78.8 & 40.8 & 89.4 & 90.3 & 41.2\\
4 & XLangMo & 78.0 & 78.4 & 61.1 & 89.4 & 90.1 & 61.4\\
\multicolumn{8}{l}{French-20K (open)} \\
1 & HLT@SUDA & 75.2 & 76.0 & 43.3 & 86.0 & 87.0 & 45.1\\
2 & XLangMo & 65.6 & 66.6 & 13.3 & 81.5 & 82.8 & 14.1\\
3 & MaskParse@Deskiñ & 65.4 & 66.6 & 24.3 & 80.9 & 82.5 & 25.8\\
4 & baseline & 48.7 & 49.6 & 2.4 & 74.0 & 75.3 & 3.2\\
5 & T\"uPa & 45.6 & 46.4 & 0 & 73.4 & 74.6 & 0
\end{tabular}
\caption{Official F1-scores for each system in each track.
Prim.: primary edges, Rem.: remote edges.
\label{tab:results}}
\end{table}

All teams found it useful to use external resources beyond those provided by the Shared Task.
Four submissions used external embeddings, MUSE \cite{conneau2017word}
in the case of {\it MaskParse@Deski\~{n}} and {\it XLangMo}, ELMo \cite{N18-1202}
in the case of {\it T\"{u}Pa},\footnote{GCN-Sem used ELMo in the closed tracks, training on the available data.}
and BERT \cite{devlin2018bert} in the case of {\it HLT@SUDA}.
Other resources included additional unlabeled data ({\it T\"{u}Pa}),
a list of multi-word expressions ({\it MaskParse@Deski\~{n}}), and the Stanford parser in the case of {\it DANGNT@UIT.VNU-HCM}.
Only {\it CUNY-PKU} used the \textit{20K} unlabeled parallel data in English and French.

A common trend for many of the systems was the use of cross-lingual projection or transfer
({\it ​MaskParse@Deski\~{n}}, {\it HLT@SUDA}, {\it T\"{u}Pa}, {\it GCN-Sem}, {\it CUNY-PKU} and {\it XLangMo}).
This was necessary for French, and was found helpful for German as well ({\it CUNY-PKU}).

\section{Results}\label{sec:results}

Table~\ref{tab:results} shows the labeled and unlabeled F1 for primary and remote
edges, for each system in each track.
Overall F1 (All) is the F1 calculated over both primary and remote edges.
Full results are available
online.\footnote{\url{http://bit.ly/semeval2019task1results}}

 \pgfplotstableread{
 category	HLT@SUDA	danielh	Davis	CUNY-PekingU	DANGNT@UIT.VNU-HCM	GCN-Sem
 Adverbial	77	73	69	72	73	70
 Center	81	77	77	77	75	71
 Connector	90	82	83	83	81	87
 Elaborator	78	74	74	74	71	74
 Function	83	80	78	81	78	80
 Ground	0	29	50	0	0	29
 Linker	86	77	74	76	76	84
 ParallelScene	76	63	66	62	53	29
 Participant	69	63	62	60	62	58
 Process	66	64	62	62	63	45
 Relator	93	92	91	91	89	91
 State	30	29	30	23	31	0
 }\englishwikiclosed
 \pgfplotstableread{
 category	HLT@SUDA	CUNY-PekingU	danielh	TuPa	XLangMo	DANGNT@UIT.VNU-HCM
 Adverbial	80	79	74	73	73	73
 Center	84	85	77	78	77	76
 Connector	93	90	84	82	85	82
 Elaborator	81	82	74	76	73	72
 Function	84	83	81	82	81	78
 Ground	0	0	0	0	0	0
 Linker	90	86	81	76	81	77
 ParallelScene	83	73	66	63	66	53
 Participant	74	72	64	64	63	61
 Process	67	66	64	65	64	62
 Relator	94	93	91	92	91	90
 State	26	34	33	31	30	29
 }\englishwikiopen
 \pgfplotstableread{
 category	HLT@SUDA	danielh	CUNY-PekingU	GCN-Sem
 Adverbial	56	53	51	52
 Center	80	76	77	73
 Connector	84	76	71	78
 Elaborator	78	74	75	74
 Function	70	70	69	65
 Ground	0	7	0	4
 Linker	72	64	62	72
 ParallelScene	61	51	48	25
 Participant	65	56	56	54
 Process	71	66	64	53
 Relator	86	84	83	84
 State	18	20	15	0
 }\englishtkclosed
 \pgfplotstableread{
 category	HLT@SUDA	CUNY-PekingU	TuPa	XLangMo	danielh
 Adverbial	60	58	54	53	54
 Center	84	81	80	79	77
 Connector	87	80	76	76	81
 Elaborator	82	81	78	76	74
 Function	71	71	74	72	71
 Ground	0	4	6	22	4
 Linker	77	76	72	70	69
 ParallelScene	68	58	54	53	54
 Participant	72	67	60	59	57
 Process	74	70	70	67	65
 Relator	87	86	84	85	84
 State	16	24	29	19	23
 }\englishtkopen
 \pgfplotstableread{
 category	HLT@SUDA	CUNY-PekingU	danielh	GCN-Sem
 Adverbial	77	72	64	69
 Center	90	87	85	81
 Connector	82	78	75	80
 Elaborator	87	83	81	82
 Function	87	85	80	85
 Ground	72	79	74	78
 Linker	88	84	76	83
 ParallelScene	78	71	33	38
 Participant	79	76	69	70
 Process	77	73	71	56
 Relator	92	90	81	92
 State	51	47	42	0
 }\germantkclosed
 \pgfplotstableread{
 category	HLT@SUDA	CUNY-PekingU	danielh	TuPa	XLangMo
 Adverbial	78	77	67	68	67
 Center	91	91	87	87	86
 Connector	82	82	82	78	80
 Elaborator	88	87	84	83	82
 Function	90	89	85	84	85
 Ground	75	80	76	69	62
 Linker	87	88	84	79	85
 ParallelScene	80	78	73	70	71
 Participant	82	80	75	73	73
 Process	80	78	74	71	71
 Relator	92	92	91	90	89
 State	54	59	46	43	27
 }\germantkopen
 \pgfplotstableread{
 category	CUNY-PekingU	HLT@SUDA	XLangMo	MaskParse@Deskin	danielh	TuPa
 Adverbial	64	50	34	46	13	11
 Center	85	83	80	76	70	68
 Connector	83	87	63	59	53	18
 Elaborator	85	79	76	71	60	57
 Function	68	61	53	53	38	24
 Ground	34	63	13	5	0	5
 Linker	83	75	63	59	36	32
 ParallelScene	65	64	46	50	16	21
 Participant	72	71	53	53	25	25
 Process	71	72	59	68	35	22
 Relator	93	88	86	84	77	69
 State	48	32	17	17	6	0
 }\frenchtkopen

 \begin{figure*}[p]
 \hspace{-4mm}
 \begin{subfigure}{\textwidth}
 \scalebox{.9}{
     \begin{tikzpicture}
     \begin{axis}[
     ybar=0pt,
     ymin=0,
     enlarge x limits={0.035},
     width=177mm,
     height=35mm,
     bar width=5pt,
     xtick=data,
     xticklabels from table={\englishwikiclosed}{category},
     xticklabel style={font=\tiny,anchor=north},
     xtick align=inside,
     xticklabel pos=left,
     yticklabels=none,
     tickwidth=0pt,
     legend style={at={(axis cs:-.5,110)},anchor=south west,font=\tiny,draw=none,legend columns=-1},
     legend cell align={left},
     nodes near coords={\scalebox{.4}{\pgfmathprintnumber[precision=2]{\pgfplotspointmeta}}},
     every node near coord/.append style={rotate=90,anchor=west}]
     \addplot[fill=green]table[x expr=\coordindex,meta=category,y=HLT@SUDA]{\englishwikiclosed};
     \addplot[fill=blue]table[x expr=\coordindex,meta=category,y=danielh]{\englishwikiclosed};
     \addplot[fill=red]table[x expr=\coordindex,meta=category,y=Davis]{\englishwikiclosed};
     \addplot[fill=orange]table[x expr=\coordindex,meta=category,y=CUNY-PekingU]{\englishwikiclosed};
     \addplot[fill=magenta]table[x expr=\coordindex,meta=category,y=DANGNT@UIT.VNU-HCM]{\englishwikiclosed};
     \addplot[fill=cyan]table[x expr=\coordindex,meta=category,y=GCN-Sem]{\englishwikiclosed};
     \legend{HLT@SUDA,baseline,Davis,CUNY-PekingU,DANGNT@UIT.VNU-HCM,GCN-Sem}
     \end{axis}
     \end{tikzpicture}}
     \caption{English Wiki (closed)}
 \end{subfigure}
 \begin{subfigure}{\textwidth}
 \scalebox{.9}{
     \begin{tikzpicture}
     \begin{axis}[
     ybar=0pt,
     ymin=0,
     enlarge x limits={0.035},
     width=177mm,
     height=35mm,
     bar width=5pt,
     xtick=data,
     xticklabels from table={\englishwikiopen}{category},
     xticklabel style={font=\tiny,anchor=north},
     xtick align=inside,
     xticklabel pos=left,
     yticklabels=none,
     tickwidth=0pt,
     legend style={at={(axis cs:-.5,110)},anchor=south west,font=\tiny,draw=none,legend columns=-1},
     legend cell align={left},
     nodes near coords={\scalebox{.4}{\pgfmathprintnumber[precision=2]{\pgfplotspointmeta}}},
     every node near coord/.append style={rotate=90,anchor=west}]
     \addplot[fill=green]table[x expr=\coordindex,meta=category,y=HLT@SUDA]{\englishwikiopen};
     \addplot[fill=orange]table[x expr=\coordindex,meta=category,y=CUNY-PekingU]{\englishwikiopen};
     \addplot[fill=blue]table[x expr=\coordindex,meta=category,y=danielh]{\englishwikiopen};
     \addplot[fill=yellow]table[x expr=\coordindex,meta=category,y=TuPa]{\englishwikiopen};
     \addplot[fill=brown]table[x expr=\coordindex,meta=category,y=XLangMo]{\englishwikiopen};
     \addplot[fill=magenta]table[x expr=\coordindex,meta=category,y=DANGNT@UIT.VNU-HCM]{\englishwikiopen};
     \legend{HLT@SUDA,CUNY-PekingU,baselines,T\"uPa,XLangMo,DANGNT@UIT.VNU-HCM}
     \end{axis}
     \end{tikzpicture}}
     \caption{English Wiki (open)}
 \end{subfigure}
 \begin{subfigure}{\textwidth}
  \scalebox{.9}{
     \begin{tikzpicture}
     \begin{axis}[
     ybar=0pt,
     ymin=0,
     enlarge x limits={0.035},
     width=177mm,
     height=35mm,
     bar width=5pt,
     xtick=data,
     xticklabels from table={\englishtkclosed}{category},
     xticklabel style={font=\tiny,anchor=north},
     xtick align=inside,
     xticklabel pos=left,
     yticklabels=none,
     tickwidth=0pt,
     legend style={at={(axis cs:-.5,100)},anchor=south west,font=\tiny,draw=none,legend columns=-1},
     legend cell align={left},
     nodes near coords={\scalebox{.4}{\pgfmathprintnumber[precision=2]{\pgfplotspointmeta}}},
     every node near coord/.append style={rotate=90,anchor=west}]
     \addplot[fill=green]table[x expr=\coordindex,meta=category,y=HLT@SUDA]{\englishtkclosed};
     \addplot[fill=blue]table[x expr=\coordindex,meta=category,y=danielh]{\englishtkclosed};
     \addplot[fill=orange]table[x expr=\coordindex,meta=category,y=CUNY-PekingU]{\englishtkclosed};
     \addplot[fill=cyan]table[x expr=\coordindex,meta=category,y=GCN-Sem]{\englishtkclosed};
     \legend{HLT@SUDA,baseline,CUNY-PekingU,GCN-Sem}
     \end{axis}
     \end{tikzpicture}}
     \caption{English 20K (closed)}
 \end{subfigure}
 \begin{subfigure}{\textwidth}
  \scalebox{.9}{
     \begin{tikzpicture}
     \begin{axis}[
     ybar=0pt,
     ymin=0,
     enlarge x limits={0.035},
     width=177mm,
     height=35mm,
     bar width=5pt,
     xtick=data,
     xticklabels from table={\englishtkopen}{category},
     xticklabel style={font=\tiny,anchor=north},
     xtick align=inside,
     xticklabel pos=left,
     yticklabels=none,
     tickwidth=0pt,
     legend style={at={(axis cs:-.5,100)},anchor=south west,font=\tiny,draw=none,legend columns=-1},
     legend cell align={left},
     nodes near coords={\scalebox{.4}{\pgfmathprintnumber[precision=2]{\pgfplotspointmeta}}},
     every node near coord/.append style={rotate=90,anchor=west}]
     \addplot[fill=green]table[x expr=\coordindex,meta=category,y=HLT@SUDA]{\englishtkopen};
     \addplot[fill=orange]table[x expr=\coordindex,meta=category,y=CUNY-PekingU]{\englishtkopen};
     \addplot[fill=yellow]table[x expr=\coordindex,meta=category,y=TuPa]{\englishtkopen};
     \addplot[fill=brown]table[x expr=\coordindex,meta=category,y=XLangMo]{\englishtkopen};
     \addplot[fill=blue]table[x expr=\coordindex,meta=category,y=danielh]{\englishtkopen};    \legend{HLT@SUDA,CUNY-PekingU,T\"uPa,XLangMo,baseline}
     \end{axis}
     \end{tikzpicture}}
     \caption{English 20K (open)}
 \end{subfigure}
 \begin{subfigure}{\textwidth}
  \scalebox{.9}{
     \begin{tikzpicture}
     \begin{axis}[
     ybar=0pt,
     ymin=0,
     enlarge x limits={0.035},
     width=177mm,
     height=35mm,
     bar width=5pt,
     xtick=data,
     xticklabels from table={\germantkclosed}{category},
     xticklabel style={font=\tiny,anchor=north},
     xtick align=inside,
     xticklabel pos=left,
     yticklabels=none,
     tickwidth=0pt,
     legend style={at={(axis cs:-.5,110)},anchor=south west,font=\tiny,draw=none,legend columns=-1},
     legend cell align={left},
     nodes near coords={\scalebox{.4}{\pgfmathprintnumber[precision=2]{\pgfplotspointmeta}}},
     every node near coord/.append style={rotate=90,anchor=west}]
     \addplot[fill=green]table[x expr=\coordindex,meta=category,y=HLT@SUDA]{\germantkclosed};
     \addplot[fill=orange]table[x expr=\coordindex,meta=category,y=CUNY-PekingU]{\germantkclosed};
     \addplot[fill=blue]table[x expr=\coordindex,meta=category,y=danielh]{\germantkclosed};
     \addplot[fill=cyan]table[x expr=\coordindex,meta=category,y=GCN-Sem]{\germantkclosed};
     \legend{HLT@SUDA,CUNY-PekingU,baseline,GCN-Sem}
     \end{axis}
     \end{tikzpicture}}
     \caption{German 20K (closed)}
 \end{subfigure}
 \begin{subfigure}{\textwidth}
  \scalebox{.9}{
     \begin{tikzpicture}
     \begin{axis}[
     ybar=0pt,
     ymin=0,
     enlarge x limits={0.035},
     width=177mm,
     height=35mm,
     bar width=5pt,
     xtick=data,
     xticklabels from table={\germantkopen}{category},
     xticklabel style={font=\tiny,anchor=north},
     xtick align=inside,
     xticklabel pos=left,
     yticklabels=none,
     tickwidth=0pt,
     legend style={at={(axis cs:-.5,110)},anchor=south west,font=\tiny,draw=none,legend columns=-1},
     legend cell align={left},
     nodes near coords={\scalebox{.4}{\pgfmathprintnumber[precision=2]{\pgfplotspointmeta}}},
     every node near coord/.append style={rotate=90,anchor=west}]
     \addplot[fill=green]table[x expr=\coordindex,meta=category,y=HLT@SUDA]{\germantkopen};
     \addplot[fill=orange]table[x expr=\coordindex,meta=category,y=CUNY-PekingU]{\germantkopen};
     \addplot[fill=blue]table[x expr=\coordindex,meta=category,y=danielh]{\germantkopen};   
     \addplot[fill=yellow]table[x expr=\coordindex,meta=category,y=TuPa]{\germantkopen};
     \addplot[fill=brown]table[x expr=\coordindex,meta=category,y=XLangMo]{\germantkopen};
     \legend{HLT@SUDA,CUNY-PekingU,baseline,T\"uPa,XLangMo}
     \end{axis}
     \end{tikzpicture}}
     \caption{German 20K (open)}
 \end{subfigure}
 \begin{subfigure}{\textwidth}
  \scalebox{.9}{
     \begin{tikzpicture}
     \begin{axis}[
     ybar=0pt,
     ymin=0,
     enlarge x limits={0.035},
     width=177mm,
     height=35mm,
     bar width=5pt,
     xtick=data,
     xticklabels from table={\frenchtkopen}{category},
     xticklabel style={font=\tiny,anchor=north},
     xtick align=inside,
     xticklabel pos=left,
     yticklabels=none,
     tickwidth=0pt,
     legend style={at={(axis cs:-.5,110)},anchor=south west,font=\tiny,draw=none,legend columns=-1},
     legend cell align={left},
     nodes near coords={\scalebox{.4}{\pgfmathprintnumber[precision=2]{\pgfplotspointmeta}}},
     every node near coord/.append style={rotate=90,anchor=west}]
     \addplot[fill=orange]table[x expr=\coordindex,meta=category,y=CUNY-PekingU]{\frenchtkopen};
     \addplot[fill=green]table[x expr=\coordindex,meta=category,y=HLT@SUDA]{\frenchtkopen};
     \addplot[fill=brown]table[x expr=\coordindex,meta=category,y=XLangMo]{\frenchtkopen};
     \addplot[fill=purple]table[x expr=\coordindex,meta=category,y=MaskParse@Deskin]{\frenchtkopen};   
     \addplot[fill=blue]table[x expr=\coordindex,meta=category,y=danielh]{\frenchtkopen};   
     \addplot[fill=yellow]table[x expr=\coordindex,meta=category,y=TuPa]{\frenchtkopen};
     \legend{CUNY-PekingU,HLT@SUDA,XLangMo,MaskParse@Deski\~n,baseline,T\"uPa}
     \end{axis}
     \end{tikzpicture}}
     \caption{French 20K (open)}
 \end{subfigure}
     \caption{Each system's labeled F1 per UCCA category in each track.
     \label{fig:fine_grained}}
 \end{figure*}
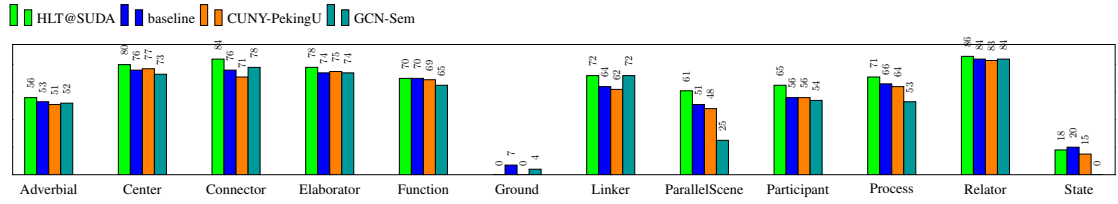
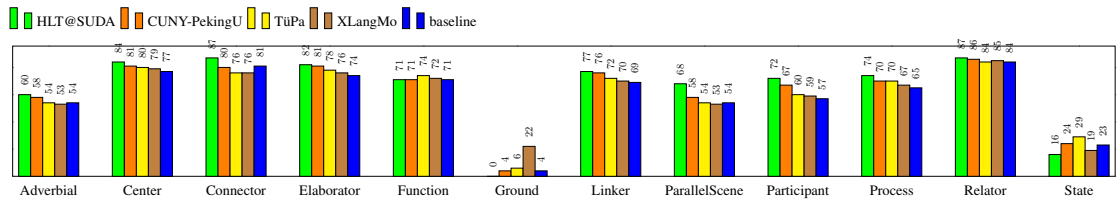
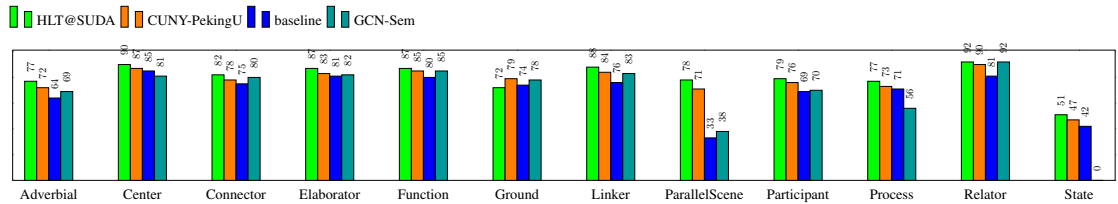
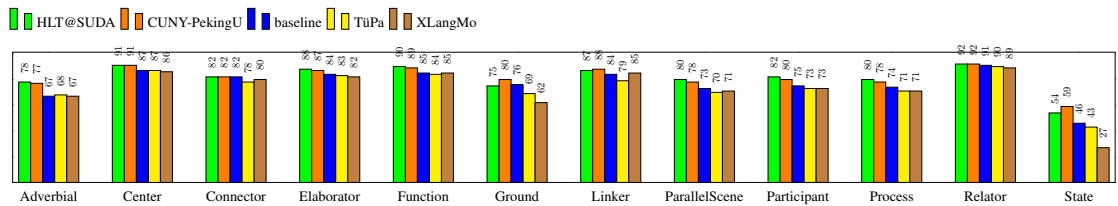
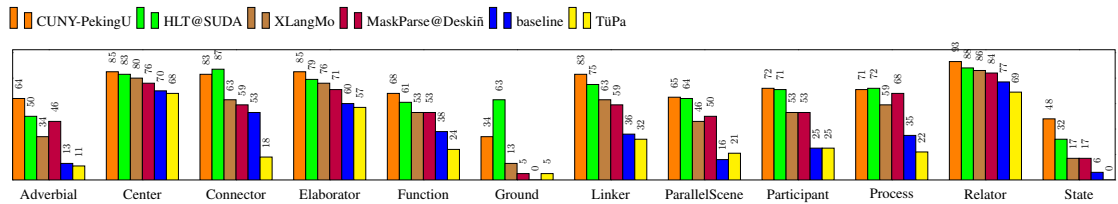

Figure~\ref{fig:fine_grained} shows the fine-grained evaluation by labeled F1
per UCCA category, for each system in each track.
While Ground edges were uniformly difficult to parse due to their sparsity
in the training data, Relators were the easiest for all systems, as they
are both common and predictable.
The Process/State distinction proved challenging, and most main relations
were identified as the more common Process category.
The winning system in most tracks (HLT@SUDA) performed better on almost
all categories. Its largest advantage was on Parallel Scenes and Linkers,
showing was especially successful at identifying Scene boundaries relative
to the other systems, which requires a good understanding of syntax.

\section{Discussion}\label{sec:discussion}

The {\it HLT@SUDA} system participated in all the tracks, obtaining the first place in the six English and German tracks and the second place in the French open track.
The system is based on the conversion of UCCA graphs into constituency trees, marking remote and discontinuous edges for recovery. The classification-based recovery of the remote edges is performed simultaneously with the constituency parsing in a multi-task learning framework. This work, which further connects between semantic and syntactic parsing, proposes a recovery mechanism that can be applied to other grammatical formalisms, enabling the conversion of a given formalism to another one for parsing. The idea of this system is inspired by the pseudo non-projective dependency parsing approach proposed by \citet{NivreNilsson2005}.

The {\it ​MaskParse@Deski\~{n}} system only participated to the French open track, focusing on cross-lingual parsing. The system uses a semantic tagger, implemented with a bidirectional GRU and a masking mechanism to recursively extract the inner semantic structures in the graph. Multilingual word embeddings are also used. 
Using the English and German training data as well as the small French trial data for training, the parser ranked fourth in the French open track with a labeled F1 score of $65.4\%$, suggesting that this new model could be useful for low-resource languages.

The {\it T\"{u}pa} system takes a transition-based approach, building on the TUPA transition system and oracle, but modifies its feature representations. Specifically, instead of representing the parser configuration using LSTMs over the partially parsed graph, stack and buffer, they use feed-forward networks with ELMo contextualized embeddings. The stack and buffer are represented by the top three items on them. For the partially parsed graph, they extract the rightmost and leftmost parents and children of the respective items, and represent them by the ELMo embedding of their form, the embedding of their dependency heads (for terminals, for non-terminals this is replaced with a learned embedding) and the embeddings of all terminal children. Results are generally on-par with the TUPA baseline, and surpass it from the out-of-domain English setting. This suggests that the TUPA architecture may be simplified, without compromising performance.

The {\it UC Davis} system participated only in the English closed track, where they achieved the second highest score, on par with TUPA. The proposed parser has an
encoder-decoder architecture, where the encoder is a simple BiLSTM encoder for each span of words. The decoder iteratively and greedily traverses the sentence, and attempts to predict span boundaries. The basic algorithm yields an unlabeled contiguous phrase-based tree, but additional modules predict the labels of the spans, discontiguous units (by joining together spans from the contiguous tree under a new node), and remote edges. The work is inspired by \citet{KitaevKlein2018}, who used similar methods for constituency parsing.

The {\it GCN-SEM} system uses a BiLSTM encoder, and predicts bi-lexical semantic dependencies
(in the SDP format) using word, token and syntactic dependency parses.
The latter is incorporated into the network with a graph convolutional network (GCN).
The team participated in the English and German closed tracks,
and were not among the highest-ranking teams. 
However, scores on the UCCA test sets converted to the bi-lexical CoNLL-U format
were rather high,
implying that the lossy conversion could be much of the reason.

The {\it CUNY-PKU} system was based on an ensemble. The ensemble included variations of TUPA parser, namely the MLP and BiLSTM models \citep{hershcovich2017a} and the BiLSTM model with an additional MLP. The system also proposes a way to aggregate the ensemble going through CKY parsing and accounting for remotes and discontinuous spans. The team participated in all tracks, including additional information in the open domain, notably synthetic data based on automatically translating annotated texts. Their system ranks first in the French open track.

The {\it DANGNT@UIT.VNU-HCM} system participated only in the English Wiki open and closed tracks. The system is based on graph transformations from dependency trees into UCCA, using heuristics to create non-terminal nodes and map the dependency relations to UCCA categories. The manual rules were developed based on the training and development data. As the system converts trees to trees and does not add reentrancies, it does not produce remote edges. While the results are not among the highest-ranking in the task, the primary labeled F1 score of 71.1\% in the English open track shows that a rule-based system on top of a leading dependency parser (the Stanford parser) can obtain reasonable results for this task.

\section{Conclusion}\label{sec:conclusion}

  The task has yielded substantial improvements to UCCA parsing in all settings. Given that the best reported results were achieved with different
  parsing and learning approaches than the baseline model TUPA (which has been the only available parser for UCCA), the task opens a variety of paths
  for future improvement. Cross-lingual transfer, which capitalizes on UCCA's tendency to be preserved in translation, was employed by a number of systems and has proven remarkably effective.
  Indeed, the high scores obtained for French parsing in a low-resource setting suggest that high quality UCCA parsing can be straightforwardly extended to additional languages, with only a minimal amount of manual labor. 

  Moreover, given the conceptual similarity between the different semantic representations \cite{abend2017state},
  it is likely the parsers developed for the shared task will directly contribute to the development of
  other semantic parsing technology. Such a contribution is facilitated by the available conversion scripts available between UCCA and other
  formats.


%

\section*{Acknowledgments}

We are deeply grateful to Dotan Dvir and the UCCA annotation team
for their diligent work on the corpora used in this shared task.

This work was supported by the Israel Science Foundation (grant No. 929/17),
and by the HUJI Cyber Security Research Center
in conjunction with the Israel National Cyber Bureau in the Prime Minister's Office.

\bibliography{references}
\bibliographystyle{acl_natbib}

\end{document}